\documentclass{article}
\usepackage{spconf,amsmath,graphicx}

\usepackage{cite}
\usepackage{algorithmic}
\usepackage{array}
\usepackage{stfloats}
\usepackage[export]{adjustbox}
\usepackage{systeme}
\usepackage[caption=false,font=footnotesize]{subfig}

\makeatletter
\def\thickhline{%
  \noalign{\ifnum0=`}\fi\hrule \@height \thickarrayrulewidth \futurelet
   \reserved@a\@xthickhline}
\def\@xthickhline{\ifx\reserved@a\thickhline
               \vskip\doublerulesep
               \vskip-\thickarrayrulewidth
             \fi
      \ifnum0=`{\fi}}
\makeatother

\newlength{\thickarrayrulewidth}
\usepackage{booktabs,makecell,multirow,tabularx}
\newcolumntype{L}{>{\RaggedRight\arraybackslash}X}
\newcolumntype{C}{>{\Centering\arraybackslash}X}

\usepackage{ragged2e}
\usepackage{amssymb}

\usepackage{tikz,xcolor,hyperref}

\definecolor{lime}{HTML}{A6CE39}
\DeclareRobustCommand{\orcidicon}{
    \begin{tikzpicture}
    \draw[lime, fill=lime] (0,0) 
    circle [radius=0.16] 
    node[white] {{\fontfamily{qag}\selectfont \tiny ID}};
    \draw[white, fill=white] (-0.0625,0.095) 
    circle [radius=0.007];
    \end{tikzpicture}
    \hspace{-2mm}
}

\foreach \x in {A, ..., Z}{\expandafter\xdef\csname orcid\x\endcsname{\noexpand\href{https://orcid.org/\csname orcidauthor\x\endcsname}
            {\noexpand\orcidicon}}
}

\title{Entropy-Based Feature Extraction for Real-Time Semantic Segmentation}
\name{Lusine~Abrahamyan$^{1,2}$\orcidD{}, Nikos~Deligiannis$^{1,2}$\orcidA{} 
\thanks{E-mail: lusine.abrahamyan@vub.be (L. Abrahamyan), ndeligia@etrovub.be (N. Deligiannis). This work was supported by the Research Foundation–Flanders (FWO) Research under Project G093817N.}}
\address{$^1$ETRO Department, Vrije Universiteit Brussel, Pleinlaan 2, B-1050 Brussels, Belgium \\
$^2$imec, Kapeldreef 75, B-3001 Leuven, Belgium}

\begin{document}
%
\maketitle
\begin{abstract}

This paper introduces an efficient patch-based computational module, coined Entropy-based Patch Encoder (EPE) module, for resource-constrained semantic segmentation. The EPE module consists of three lightweight fully-convolutional encoders, each extracting features from image patches with a different amount of entropy. Patches with high entropy are being processed by the encoder with the largest number of parameters, patches with moderate entropy are processed by the encoder with a moderate number of parameters, and patches with low entropy are processed by the smallest encoder. The intuition behind the module is the following: as patches with high entropy contain more information, they need an encoder with more parameters, unlike low entropy patches, which can be processed using a small encoder. Consequently, processing part of the patches via the smaller encoder can significantly reduce the computational cost of the module. Experiments show that EPE can boost the performance of existing real-time semantic segmentation models with a slight increase in the computational cost. Specifically, EPE increases the mIOU performance of DFANet A by $0.9\%$ with only $1.2\%$ increase in the number of parameters and the mIOU performance of EDANet by $1\%$ with $10\%$ increase of the model parameters. 

\end{abstract}

\begin{figure}[ht]
\begin{center}
   \includegraphics[width=1.0\linewidth]{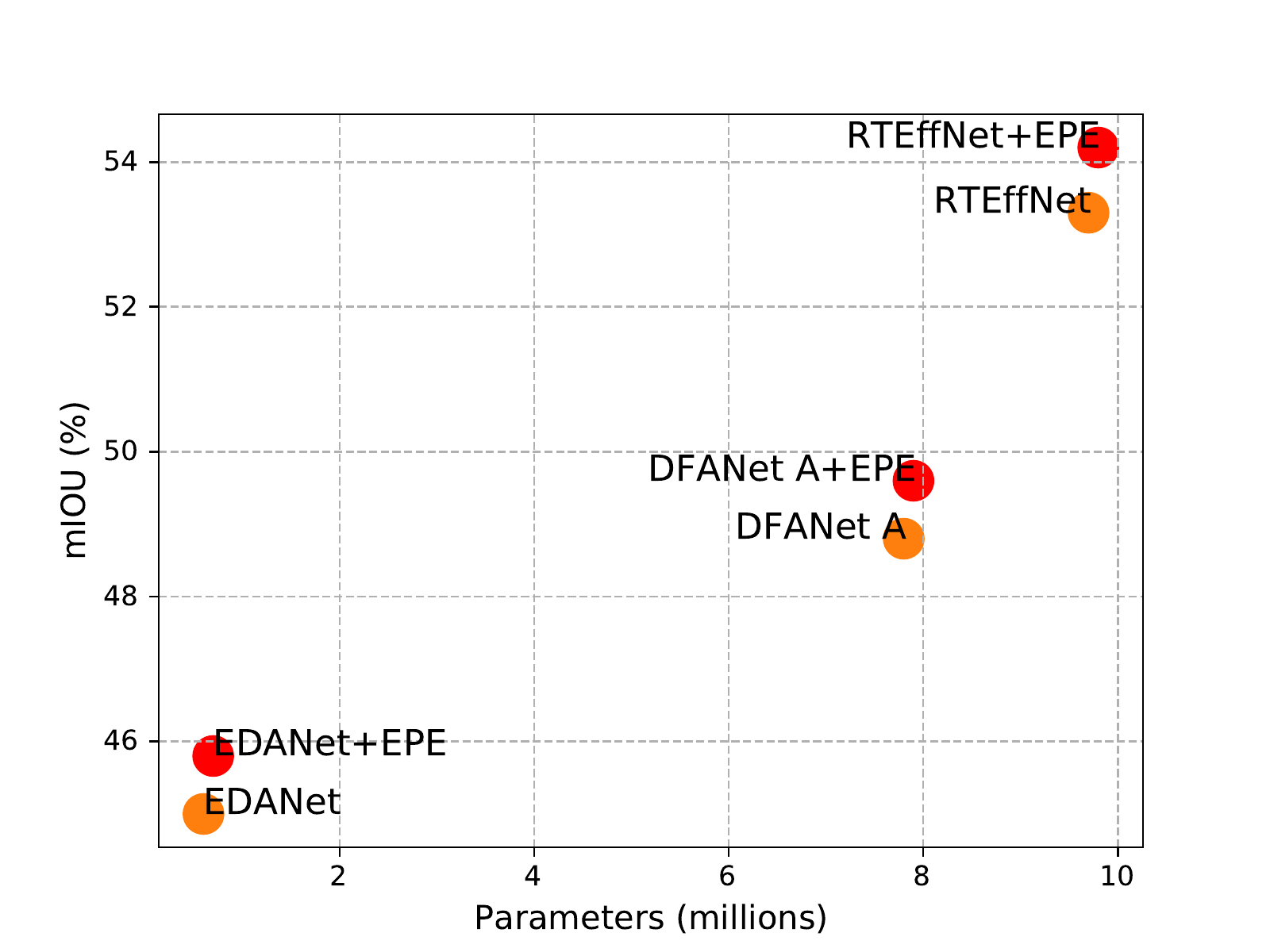}
\end{center}
   \caption{Number of parameters and mIOU on the CamVid dataset for the models trained with and without EPE module. }
\label{fig:camvid_results}
\end{figure}

\begin{keywords}
Real-time semantic segmentation, deep learning, neural network.
\end{keywords}
\section{Introduction}
\label{sec:intro}

Semantic segmentation is a fundamental computer vision task, which aims to assign a dense label to each pixel in the image. In recent years, the development of convolutional neural networks (CNNs) has led to substantial progress in high-quality semantic segmentation~\cite{nirkin2021hyperseg, zhao2017pyramid, chen2018atrous}. High accuracy semantic segmentation is mainly achieved by large models of considerable depth and width, which have a huge number of parameters and high latency. However, with the growing interest in applications like autonomous driving, robotics, and intelligent surveillance, the demand for real-time semantic segmentation is drastically rising. Thus, developing efficient architectures for real-time semantic segmentation is essential.

Several studies have focused on designing resource constraint architectures and modules for semantic segmentation. The authors of EDANet~\cite{edanet} proposed using asymmetric convolutions to reduce the number of parameters of the model. In the DFANet~\cite{li2019dfanet} model, features at different stages are combined in the processing path of the network to enhance feature representation. Several studies suggest using pretrained backbones~\cite{chen2018atrous, nirkin2021hyperseg} to boost the model performance without increasing the number of parameters and utilizing a cost-efficient replacement of the conventional convolution, i.e.,, the depthwise separable convolution~\cite{chollet2017xception}, to reduce computational cost. 

In this work, we propose to complement existing architectures for real-time semantic segmentation with a novel entropy-based patch encoder module that boosts performance with only a slight increase in the number of parameters. Specifically, we propose to use entropy as a measure of informativeness for the patches extracted from the input image and process them using different encoders. As a result, patches with high entropy will be processed by an encoder with more trainable parameters and patches with lower entropy with a smaller encoder. Our experimental results indicate that the proposed EPE module improves notably the performance of existing and new models for real-time semantic segmentation, inclduing EDANet~\cite{edanet}, DFANet~\cite{li2019dfanet} and a model based on EfficientNet~\cite{tan2019efficientnet} that is designed by us.

The rest of the paper is organized as follows. Section~\ref{sec:RelatedWork} reviews the related work and Section~\ref{sec:epe_module} presents the proposed EPE module. Experimental results are presented in Section~\ref{sec:experiments}, and conclusions are drawn in Section~\ref{sec:conclusion}.

\section{Related Work}
\label{sec:RelatedWork}
\subsection{Real-Time Semantic Segmentation}

Real-time semantic segmentation algorithms aim to generate high-quality predictions under recourse-constraint conditions. The authors of EDANet~\cite{edanet} proposed decomposing a standard 2D convolution into two 1D convolutions, i.e., an  $n\times n$ convolutional kernel is factorized into two kernels, $n\times 1$ and $1\times n$. This approach helps reducing the number of parameters with negligible performance degradation. In DFANet~\cite{li2019dfanet}, the authors utilized the pretrained Xception~\cite{chollet2017xception} model, consisting of depthwise separable convolutions, to increase the performance. Further, to capture the contextual information at multiple scales cost-efficiently, the authors of~\cite{chen2018atrous} integrated atrous separable convolutions into the Xception backbone.

To further increase the performance of real-time semantic segmentation networks, we propose a lightweight computational module, namely, Entropy-based Patch Encoder (EPE), that can be easily integrated into existing architectures.

\subsection{Patch-based Image Processing}
The authors of the ConvMixer~\cite{patchesallyouneed} model showed that a simple architecture that directly operates on patches can outperform models such as ResNet~\cite{resnet} and Vision Transformer (ViT)~\cite{dosovitskiy2020vit} with the similar number of parameters. ConvMixer uses patch embeddings to preserve locality and  applies multiple fully-convolutional blocks consisting of large-kernel depthwise convolutions~\cite{chollet2017xception} followed by pointwise convolutions~\cite{mobilenets} that feed a global pooling operation and then a linear classifier. L. Melas-Kyriazi~\cite{melas2021resmlp} proposed a similar patch-based encoding approach where patches are processed by a combination of MLP-based cross-channel layers and convolutions.

Driven by the success of the presented works, we propose to not only process the input using a patch-based approach but also extract features from the patches using different encoders. The more informative the patch is, the higher the number of parameters in the encoder. Feature maps extracted through the multiple encoders are further used to enrich the features of the main network and boost the performance.

\begin{figure*}[ht]
\begin{center}
  \includegraphics[width=1.0\textwidth]{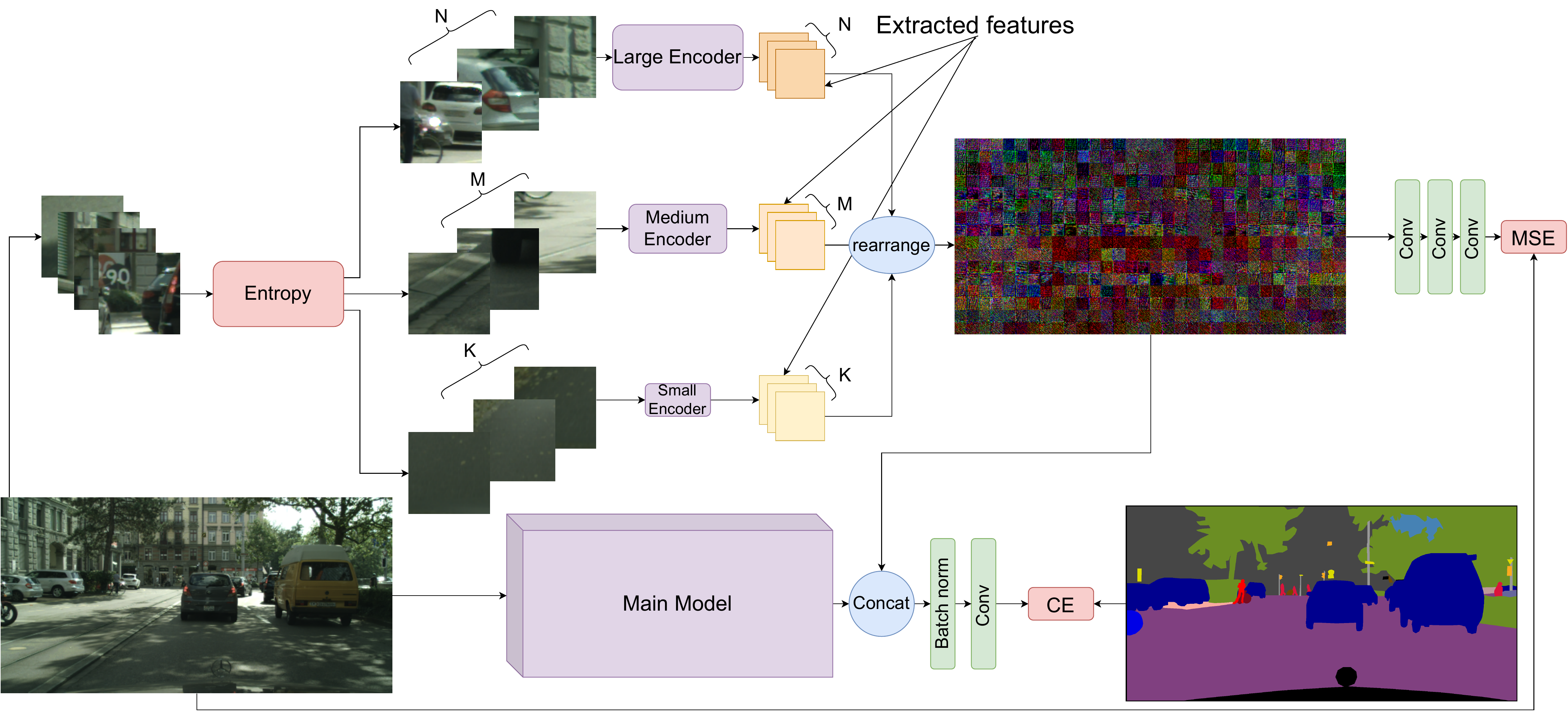}
\end{center}
  \caption{The architecture of the proposed EPE module: the Main Model can be any neural network for semantic segmentation. CE and MSE correspond to the Cross-Entropy and the mean squared-error (MSE) loss, respectively.}
\label{fig:epe_module}
\end{figure*}

\section{Entropy-based Patch Encoder}
\label{sec:epe_module}

In this work, we propose a lightweight patch-based feature extraction module that can increase the performance of real-time semantic segmentation models with only a slight increase in the number of parameters. Our module (see Fig.~\ref{fig:epe_module}) consists of three encoders with a different number of trainable parameters. The encoder with the largest number of parameters is designed to process the patches with the highest entropy. Further, the remaining two encoders with moderate and small number of parameters are responsible for processing patches with moderate and low entropy, respectively. 

The entropy of a patch is calculated as follows: an input RGB image tensor of size $c \times h \times w$ $(c=3)$---where $h$ and $w$ are the image height and width, respectively---is converted into a grayscale representation and then unfolded into $n \times n$ non-overlapping patches, thereby forming a matrix with dimensions $n^2 \times \frac{w \cdot h}{n^{2}}$, whose columns are the vectorised patches. Per column $i$ (corresponding to the $i$-th patch), we apply the kernel density estimation (KDE) method on the 32-bit uniformly quantized  values leading to the calculation of the probability density function,
\begin{equation}
    \Tilde{p}_{i}(x) = \frac{1}{n^{2}h} \sum_{j=1}^{n^{2}} K\Big( \frac{x - X_{j}}{h}\Big ),
\end{equation}
where $n$ is the size of the patch, $X_{j}$ is the $i$-th quantized element in the patch,  $K$ is the kernel (a non-negative function) and $h > 0$ is a smoothing parameter.
Then, using the probability density function,  the entropy of the $i$-th patch is estimated as
\begin{equation}
    H_{i}(x) = - \Tilde{p}_{i}(x) \log (\Tilde{p}_{i}(x)).
\end{equation}
When the entropy of each patch is estimated, the patches are being divided into 3 groups: (i) a group consisting of $20\%$ patches with the highest entropy values, (ii) a group containing $40\%$ of patches with moderate entropy values, and (iii) a group with the remaining $40\%$ of patches. Our empirical evaluation suggests that the percentages presented above provide the best trade-off between computational complexity and performance.

Further, these groups of patches form three tensors with shape $b\times g_{k} \times n \times n$, where $g_{k}$ is the number of patches in the $k$-th ($k\in [0,2]$) group and $b$ is the number of samples in the batch, are being fed to the corresponding encoders. The three encoders are realised by fully convolutional neural networks with different computational complexities. By using encoders with different computational complexities for different patches, we are able to provide an additional increase to the performance with only a subtle increase in the number of parameters. In the next step, the outputs of the encoders with  shape $b\times g_{k} \times n \times n$ are being folded back into the shape of the input grayscale image $b \times 1 \times h \times w$ in order to be concatenated with the main model for semantic segmentation. In addition, after the concatination, batch normalization is applied before proceeding to the final convolution.

\subsection{Encoder Architecture}
Each fully convolutional encoder in the EPE module (see Fig.~\ref{fig:epe_module}) consists of six sequentially stacked residual blocks~\cite{resnet}. The residual blocks of each encoder differ from each other in the number of filters in the convolutional layers: the encoder responsible for processing patches with high entropy values (a.k.a., \textit{Large Encoder}) has 16 filters, the encoder that extracts features from  patches with moderate entropy (\textit{Medium Encoder}) has 8 filters and the encoder for patches with low entropy (\textit{Small Encoder}) has 4 filters. Further, following the practice in~\cite{li2019dfanet, chen2018atrous}, the traditional convolutions are replaced with the separable depthwise convolutions~\cite{chollet2017xception}. Within a training, three convolutional layers are added at the end, and the mean squared-error (MSE) is calculated between the output of the last convolutional layer and the input RGB image.

\begin{figure*}[ht!]
\begin{center}
  \includegraphics[width=1.0\linewidth]{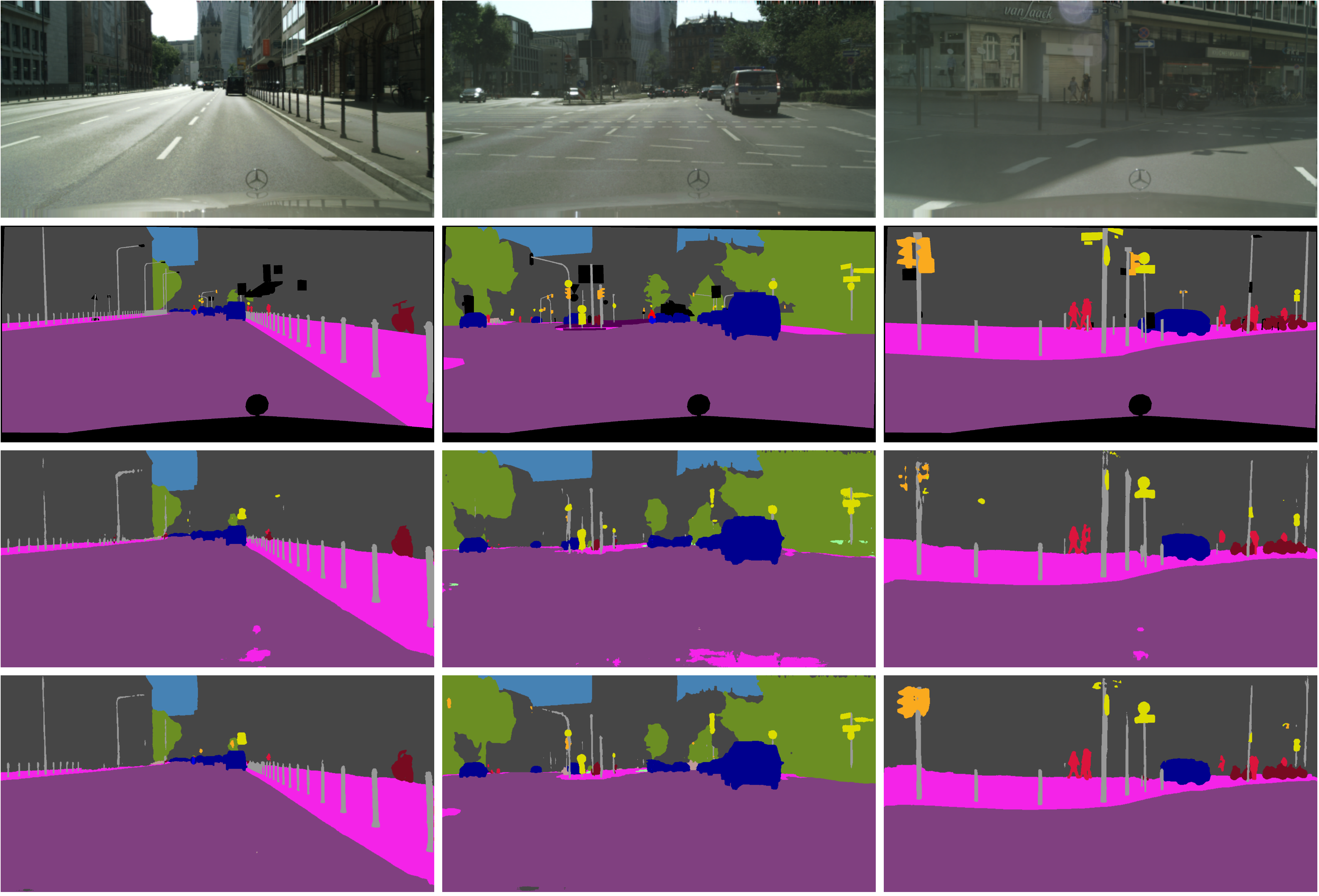}
\end{center}
  \caption{Visual results of the EPE module on the Cityscapes validation set. Input images (upper row); ground-truth segmentation masks (2nd row); masks produced by the RTEffNet model (3rd row); masks produced by the RTEffNet+EPE model (4th row).}
\label{fig:img_comparison}
\end{figure*}

\section{Experiments}
\label{sec:experiments}

We present empirical results to demonstrate the efficiency of our EPE module. All experiments were performed on a single machine with 2 GeForce RTX 2080 Ti GPUs. We used PyTorch~\cite{pytorch} as a machine learning framework. Moreover, within all experiments, we used a fixed seed and only deterministic versions of the algorithms to ensure the reproducibility of the results.

\subsection{Real-Time Semantic Segmentation}
\textbf{Datasets and Evaluation Metrics}: We evaluate the proposed EPE module's performance on the Cityscapes~\cite{Cordts2016Cityscapes} and CamVid~\cite{camvid} datasets.  Cityscapes is an urban street scene dataset
that contains $19$ object classes. It consists of $5000$ fine-annotated high-resolution images with a spatial dimension of $1024 \times 2048$ pixels, which are split into three sets: $2975$ images for training, $500$ images for validation, and $1525$ images for testing. CamVid contains images with $32$ object classes extracted from video sequences
with resolution up to $960 \times 720$ pixels. It contains $367$ images for training and $101$ for validation.
All the reported accuracy results are measured in the mean intersection over union (mIoU)~\cite{jaccard1912distribution} metric.

\textbf{Models and Training}: We choose two well-performing semantic segmentation models, i.e., EDANet~\cite{edanet} and DFANet A~\cite{li2019dfanet}, and one custom U-type model, coined RTEffNet, with the encoder realised in the form of EfficientNet-b6~\cite{tan2019efficientnet} and the decoder consisting of four bilinear upsampling layers and four residual blocks~\cite{resnet}. For all models, we first conduct trainings without the EPE module and compare the results with the corresponding models trained with the EPE module. Further within all trainings, we used the following setup: the Adam optimizer~\cite{kingma2014adam} with weight decay $1e-4$, poly learning rate policy, where the learning rate is multiplied by \texttt{(1- iter / max\char`_iter)$\times$power} with $\texttt{power}=0.9$ and initial learning rate $1e-3$.

We adopt the spatial resolution of images equal to $512 \times 1024$  and $768 \times 576$ pixels for the trainings on Cityscapes and CamVid, respectively. Further, we used random horizontal flip, random scale in a range of $[0.5, 2.0]$, and random rotation in the range of $[-10, 10]$ as data augmentation on the training phase. For our EPE module in all of our experiments we adopt a patch size of $n=32$.

\textbf{Experimental  Results}: The results presented in Table~\ref{table:cityscapes_results} and Figure~\ref{fig:camvid_results} indicate that for all tested models the proposed EPE module can boost the mIOU by $0.6\%-1.0\%$. Specifically, on Cityscapes dtaset, it increases the mIOU of the DFANet A model by $0.9\%$ with only $1.2\%$ increase in the number of parameters, and the mIOU of RTEffNet by $2.4 \%$ with $0.8\%$ increase in the number of trainable parameters.

\begin{table}
\caption{The evaluation results on the Cityscapes dataset.}
\begin{center}
\scalebox{1.0}{
    \begin{tabular}{l||c|c }
    \thickhline
     Model                  & Params        & mIOU ($\%$)\\
      \thickhline
     EDANet                 &  0.69M        &  60.5 \\
     EDANet + EPE           &  0.76M        &  61.5 \\
     \hline
     DFANet A               & 7.8M          & 67.2   \\
     DFANet A + EPE         & 7.9M          & 68.1   \\
     \hline
     RTEffNet                 & 9.7M          & 65.2  \\
     RTEffNet + EPE           & 9.9M          & 66.0  \\
      \thickhline
\end{tabular}}
\end{center}
\label{table:cityscapes_results}
\end{table}

\section{Conclusion}
\label{sec:conclusion}
In this paper, we have proposed a new plug-in entropy-based patch encoder module for the real-time semantic segmentation task in order to provide additional boost to the performance of the models. Evaluation performed on two benchmark datasets, Cityscapes~\cite{Cordts2016Cityscapes} and CamVid~\cite{camvid}, using well-performing real-time semantic segmentation models, shows that the proposed EPE module can systematically improve the mIOU with only slight increase in the number of trainable parameters.

\bibliographystyle{IEEEbib}
\bibliography{egbib}

\end{document}